\documentclass[wcp]{jmlr}

\usepackage{longtable}
\usepackage{booktabs}
\usepackage{array}
\usepackage{amsfonts}
\usepackage{nicefrac}
\usepackage[expansion=false]{microtype}

\usepackage{lineno}

\makeatletter
\let\Ginclude@graphics\@org@Ginclude@graphics
\makeatother

\newcommand{\up}{{\color{red}$\uparrow$}}
\newcommand{\dn}{{\color{blue}$\downarrow$}}

\jmlryear{2026}
\jmlrworkshop{OE-Agent @ ACML 2026}

\title[Learning When to Ask]{Learning \emph{When} to Ask: Advisor-Aware Post-Training for SLM Agents}

\author{\Name{Aasheesh Singh} \Email{aasheesh.singh@quantiphi.com}\\
 \Name{Vishal Vaddina}\\
 \addr Phi Labs, Quantiphi}

\begin{document}

\makeatletter
\let \@jmlrpages \@empty
\makeatother

\maketitle

\begin{abstract}
Small language models (SLMs) are increasingly attractive for cheap, high-throughput agentic
deployment, but they remain weak at multi-step reasoning and tool use that long-horizon
tasks demand. We study how to post-train an SLM to
\emph{seek advice} from a stronger model and act on that advice itself, following an
advisor--executor pattern. Out of the box, this pattern fails across open- and
closed-source small-model families: wiring an SLM to an advisor is a
harness-design matter that is not enough on its own. A post-training recipe is needed to teach
the model \emph{when} to ask, in a cost-aware fashion, and \emph{how} to follow advice across
steps. Our method, \textbf{ADAPT} (\textbf{AD}visor-\textbf{A}ware \textbf{P}ost-\textbf{T}raining),
is a multi-stage recipe that instills this behavior, learning entirely from the policy's own
rollouts supervised by the environment's outcome, with no learned reward model or human
annotation. Across three domains, compared to an untrained harness baseline, ADAPT
improves success rate in embodied text reasoning (TextWorld) $24.5\!\to\!77.3\%$, code
generation (LiveCodeBench) $21.8\!\to\!56.0\%$, and deep web search (GAIA)
$17.5\!\to\!49.3\%$, while querying the advisor sparingly (${\sim}1.0$--$2.1$ calls per task). On the two long-horizon
domains it closes $70$--$85\%$ of the gap to an oracle that perfectly times when to ask for advice,
and on code generation it exceeds that oracle.
The gains persist when the advisor is replaced by one unseen in training, showing
that the learned help-seeking policy is robust to a changing advisor rather than tied to the one
it was trained with.
\end{abstract}

\section{Introduction}

Language-model agents that plan and act over long horizons are increasingly capable, but the
strongest models are costly to serve at the throughput such loops demand. Small language models
(SLMs) invert this trade-off, being cheaper and faster at the price of weaker long-horizon
reasoning and tool use. A practical way to recover quality, proposed in the recent Opus~5 release
by Anthropic~\citep{anthropic2026advisor}, is to give the smaller model access to a larger advisor
and let it ask for advice as a tool call. Knowing \emph{when} that help is worth requesting and
\emph{how} to use it once received is itself a skill, and we argue empirically that it is not
innate to the harness, especially in small models.

This turns out to be a genuine gap. Equipping a range of off-the-shelf policies, open-weight and
frontier ``lite''-class alike, with the same advisor and request tool, we find that none ask well
(Table~\ref{tab:motivation}): some almost never ask even when stuck, others ask constantly without
turning advice into progress, and many ask too late to recover. Prompting changes how often a model
asks but not how well-timed or well-used those asks are. The deficiency is one of calibration and
grounding rather than exposure, and it persists across model scale and task, which makes
\emph{when to ask} a target for training rather than prompting. Closing this gap is what our
method sets out to do, and our main contributions are as follows:

\begin{table}[t]
  \centering
  \caption{Out-of-the-box, the advisor--executor pattern is miscalibrated everywhere.
  Each model carries an \texttt{ask\_advisor} tool; the advisor is
  Gemini~3.7~Flash for every model. \emph{SR\%\,(s$\to$a)} is success rate solo vs.\ advised; \emph{ask/ep}
  is mean advisor calls per episode (\up\,over- / \dn\,under-asks relative to the oracle
  timing policy); \emph{Orc\%} is the oracle-ceiling SR (if asked at the optimal steps). Simply plugging in
  an advisor as a tool is not enough, especially for small models, which rarely ask optimally;
  this behavior has to be learned through post-training.}
  \label{tab:motivation}
  \footnotesize
  \setlength{\tabcolsep}{3pt}
  \begin{tabular}{@{}l ccc ccc ccc@{}}
    \toprule
    & \multicolumn{3}{c}{\textbf{TextWorld}} & \multicolumn{3}{c}{\textbf{LCB-stdin}}
    & \multicolumn{3}{c}{\textbf{GAIA-Text}} \\
    \cmidrule(lr){2-4}\cmidrule(lr){5-7}\cmidrule(lr){8-10}
    Model & SR\%\,{\scriptsize(s$\to$a)} & ask/ep & Orc\% & SR\%\,{\scriptsize(s$\to$a)} & ask/ep & Orc\% & SR\%\,{\scriptsize(s$\to$a)} & ask/ep & Orc\% \\
    \midrule
    \multicolumn{10}{l}{\emph{Open-weight}}\\
    Qwen3-8B          & $21\!\to\!28$ & 0.12\,\dn & 86 & $22\!\to\!36$ & 0.88\,\dn & 52 & $15\!\to\!19$ & 0.11\,\dn & 62\\
    Llama-3.1-8B      & $9\!\to\!19$ & 0.15\,\dn & 54 & $15\!\to\!23$ & 0.72\,\dn & 46 & $4.2\!\to\!11$ & 0.80\,\dn & 28  \\
    \midrule
    \multicolumn{10}{l}{\emph{Closed}}\\
    Gemini 3.5 Flash-Lite & $69\!\to\!76$ & 3.2\,\up & 89 & $51\!\to\!68$ & 2.25\,\up & 74 & $54\!\to\!65$ & 4.14\,\up &  71\\
    GPT-5.4 nano          & $28\!\to\!31$ & 0.4\,\dn & 62 & $62\!\to\!60$ & 0.1\,\dn & 75 & $26\!\to\!29$ & 0.25\,\dn & 45 \\
    Claude Haiku 4.5      & $51\!\to\!58$ & 0.1\,\dn & 78 & $41\!\to\!47$ & 0.7\,\dn & 63 & $39\!\to\!42$ & 0.75\,\dn & 52 \\
    \bottomrule
  \end{tabular}
\end{table}

\begin{enumerate}\setlength{\itemsep}{1pt}\setlength{\parskip}{0pt}
  \item \textbf{Characterization of advisor-seeking.} We show that off-the-shelf agents, small
    and frontier alike, do not reliably know \emph{when} to ask a stronger model for help across
    three agentic benchmarks, and that prompting changes how often they ask but not how well
    (Section~\ref{sec:results}).
  \item \textbf{The ADAPT recipe.} We introduce \textbf{AD}visor-\textbf{A}ware
    \textbf{P}ost-\textbf{T}raining, a three-stage recipe in which a supervised warm-start installs basic advice-seeking,
    preference optimization over counterfactually mined pairs sharpens \emph{when} to ask,
    \emph{when not} to ask, and \emph{how} to act on advice, and a cost-constrained RL stage prices
    each ask. The recipe learns entirely from the policy's own environment-verified rollouts, with
    no learned reward model or human annotation (Section~\ref{sec:method}).
  \item \textbf{Gains across domains, robust to a changing advisor.} ADAPT is best or near-best on
    all three domains, more than tripling untrained success on TextWorld ($24.5\%\!\to\!77.3\%$)
    while querying the advisor sparingly, and its gains persist when the advisor is replaced by one
    unseen in training (Section~\ref{sec:results}).
\end{enumerate}

\section{Related Work}
\label{sec:related}

\paragraph{Routing versus advice-seeking.}
The dominant way to combine a small model with a stronger one is to route between them, either by
cascading, where the cheap model runs first and escalates when a confidence or verifier signal is
low~\citep{frugalgpt2023}, or with learned routers that predict up front which model should answer
each query~\citep{routellm2024}. \citet{r2v2026} bring this to the agentic, per-step setting,
sampling candidate actions from a frozen SLM, scoring them with a learned verifier, and routing to a
teacher. All of these \emph{delegate}, passing control to the larger model, whereas ADAPT keeps the
SLM in control as it asks for advice and executes the plan itself, with the help-seeking behavior
learned in its own weights. Of these methods \citet{r2v2026} is closest to ours, but its code,
checkpoints, and evaluation splits are not public, so we contrast the two designs analytically
rather than benchmarking head to head.

\paragraph{Critical-step preference optimization and counterfactual attribution.}
\citet{cso2026} mine DPO pairs at outcome-critical steps, proposing alternatives with a teacher,
scoring them with a learned process reward model, and verifying the best by replaying the
environment, while \citet{stepdpo2024} localize preference optimization to individual reasoning
steps. We adopt the critical-step idea and the environment-verified replay but drop the reward
model, since each of our environments already returns a verifiable outcome. The replay itself is
the same-policy counterfactual that \citet{car2026} formalize for debugging agent failures; we
borrow its discipline to decide which step flips an outcome, here to \emph{mine} training pairs.

\paragraph{Learning to ask, and foundations.}
Closest in spirit, \citet{ramrakhya2025ask} train embodied multimodal agents to request help
through reinforcement learning; we share the goal but install it by preference optimization as a
native tool call, and add cost-awareness so that the policy asks only when help is worth its price.
Our training data come entirely from the policy's own rollouts, in the spirit of bootstrapping from
self-generated traces~\citep{star2022}, and we build on ReAct-style tool use~\citep{react2023},
DPO~\citep{dpo2023}, LoRA~\citep{lora2022}, and GRPO~\citep{grpo2024}.

\section{Method: ADAPT}
\label{sec:method}

\begin{figure}[t]
  \centering
  \includegraphics[width=0.85\textwidth]{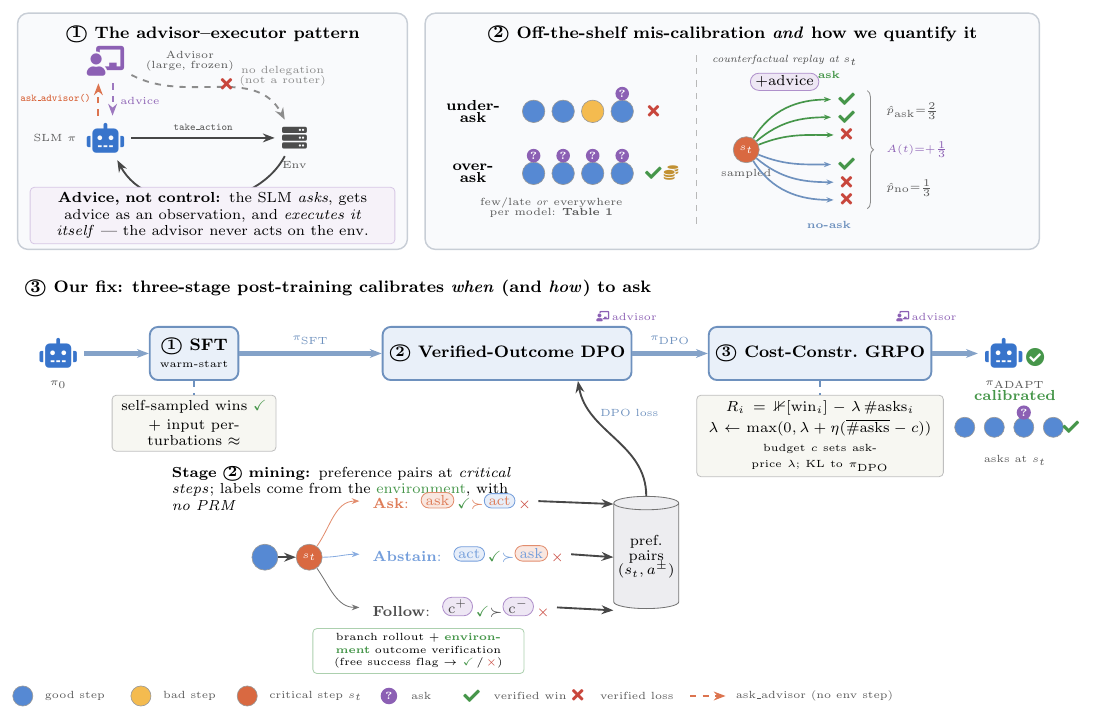}
  \caption{\textbf{ADAPT in one view.} (\textcircled{\scriptsize 1}) The SLM asks the advisor
    and executes the returned advice itself, so the advisor never acts on the environment.
    (\textcircled{\scriptsize 2}) Off-the-shelf small models either under-ask or over-ask
    (Table~\ref{tab:motivation}). (\textcircled{\scriptsize 3}) ADAPT post-trains the SLM in three
    stages: an SFT warm-start, DPO on Ask, Abstain, and Follow pairs mined at critical steps and
    labeled by the environment outcome (Section~\ref{sec:mining}), and cost-constrained GRPO that
    prices each ask.}
  \label{fig:pipeline}
\end{figure}

\subsection{Problem setup and method overview}
\label{sec:interface}

The policy $\pi_\theta$ is an SLM that acts in a text environment through tool calls in the
ordinary decode loop, producing a chain-of-thought before each action. On top of each
harness's native action tools we expose a single additional tool,
$\texttt{ask\_advisor}(\textit{request})$, whose one argument is a short natural-language
request that the policy writes itself to describe where it is stuck. We borrow the
advisor--executor formulation of \citet{anthropic2026advisor}: when the tool is called, the
advisor reads the full trajectory so far and returns strategic guidance, which we hand back
to the policy as an ordinary observation without consuming an environment step. The advisor only
ever produces advice; it never
acts on the environment, so control of the episode stays with the SLM throughout.

ADAPT teaches the policy to spend these advisor calls well, asking when it helps and refraining
when it does not, through a three-stage pipeline (Figure~\ref{fig:pipeline}), each stage adding a
distinct competence: SFT installs the ability to seek and use advice; DPO calibrates \emph{when} to ask
(and when not) and how to follow advice; and cost-constrained RL prices each ask to moderate
frequency without giving back success. All three learn from the policy's own rollouts, labeled
only by the environment's outcome.

\subsection{Stage 1: supervised warm-start}
\label{sec:sft}

We first fine-tune the base SLM on its own successful trajectories, collected by rolling
out the policy with the \texttt{ask\_advisor} tool available and keeping the episodes that
solve the task. The retained set deliberately mixes episodes that used the advisor with
episodes that did not, so the warm-start installs both a well-formed \texttt{ask\_advisor}
call and competent unaided behavior. To broaden coverage, we augment the traces with light
input perturbations~\citep{r2v2026}, namely paraphrased observations and perturbed
surface features, so the same decision is seen under several textual variants. This stage installs the
behavior but not its calibration: the warm-started policy can ask and act on advice, but does
not yet know \emph{when} either is right, which is the job of Stage 2.

\subsection{Stage 2: counterfactual preference mining and DPO}
\label{sec:mining}

\paragraph{Where to probe.} Estimating the effect of asking at every step is prohibitively
expensive, so for each trajectory we probe a subset of candidate steps, one quarter of the step
horizon, sampled uniformly at random along its length. On TextWorld, for instance, this amounts
to $5$ of the $20$ steps in the horizon. Where a trajectory exposes them, we further bias this
selection toward likely-critical steps using lightweight signals: syntax errors, tool-call
failures or timeouts, and loop detection, each a marker that the policy is plausibly stuck.

\paragraph{Counterfactual replay.} At a candidate step $t$ we roll the policy forward to the
end of the episode $K$ times under two interventions (panel~2 of Figure~\ref{fig:pipeline}): one in which the policy asks at $t$
(the advisor's advice is injected as an observation and the episode continues), and one in
which it acts without asking. Scoring both sets of continuations by the environment's win
flag gives two estimates, $\hat p_{\text{ask}}(t)=P(\text{win}\mid\text{ask at }t)$ and
$\hat p_{\text{noask}}(t)=P(\text{win}\mid\text{act at }t)$, whose difference is the
\emph{ask-advantage}
\begin{equation}
  A(t)=\hat p_{\text{ask}}(t)-\hat p_{\text{noask}}(t),
\end{equation}
a measure of how much asking at $t$ changes the chance of success. The same sweep yields the oracle-timing ceiling reported in our motivating
analysis (Table~\ref{tab:motivation}): the success rate attainable if the policy asked at
its single most valuable step.

\paragraph{From estimates to preference pairs.} We turn these estimates into three kinds of
preference pairs, one for each competence the policy must acquire (summarized in
Table~\ref{tab:pairs}, Appendix~\ref{app:pairs}), and keep a step only when its estimates clear a
margin, which discards the noisy near-ties in the middle of the ask-advantage range:
\begin{itemize}\itemsep2pt
  \item \textbf{Ask} pairs teach \emph{when to ask}. When asking reliably rescues a step the
    policy would otherwise fail, $\hat p_{\text{ask}}(t)>0.6$ and $A(t)>0.3$, the chosen branch
    calls \texttt{ask\_advisor} and the rejected branch takes the direct action.
  \item \textbf{Abstain} pairs teach \emph{when not to ask}. When the policy already solves the
    step and asking adds nothing, $\hat p_{\text{noask}}(t)>0.6$ and $A(t)\le 0$, the chosen
    branch acts directly and the rejected branch asks. These pairs give the policy a reason to
    hold back and keep the ask-rate down.
  \item \textbf{Follow} pairs teach \emph{how to act on advice}. The chosen branch follows the
    advice through to a win, while the rejected branch receives the same advice but drifts to a
    loss. These pairs score the whole continuation to the trace end, backpropagating over all
    steps that succeed the advice step, since acting on advice plays out across several turns.
\end{itemize}

\paragraph{DPO training.} We optimize the retained pairs with DPO~\citep{dpo2023}, taking the
Stage-1 checkpoint as the frozen reference $\pi_\text{ref}$ and adapting with LoRA~\citep{lora2022}
to limit drift from the warm-started behavior. For Ask and Abstain pairs we apply one-step DPO on
the immediate step succeeding the current critical step, scoring only the policy's decision, the
\texttt{ask\_advisor} call or the direct action, whereas Follow pairs use a multi-step DPO loss
over the continuation until the end of the trajectory. Injected advice is an observation rather
than a policy action, so advice tokens are masked as well.

\subsection{Stage 3: cost-constrained GRPO}
\label{sec:grpo}

DPO has no notion of what an ask costs: its ask-rate is fixed by the balance of the mined
pairs, and in practice the DPO policy over-asks, buying success with more advisor calls than a
deployment budget allows. Stage 3 makes cost a quantity the user controls. We cast it as a
constrained problem, maximizing success subject to an ask budget:
\begin{equation}
  \max_{\pi}\ \mathbb{E}[\text{win}]\quad\text{s.t.}\quad \mathbb{E}[n^{\text{ask}}]\le c,
\end{equation}
where $\text{win}\in\{0,1\}$ is the environment's success flag and $n^{\text{ask}}$ is the
number of \texttt{ask\_advisor} calls in an episode, an \emph{observed} count, the quantity
being budgeted. The budget $c$, the \emph{maximum} advisor calls per episode, is the only knob
the user sets; the method then finds its own operating ask-rate, which typically settles well
below $c$.

Relaxing the constraint with a multiplier $\lambda\ge0$ gives the Lagrangian
$\mathcal{L}=\mathbb{E}[\text{win}]-\lambda(\mathbb{E}[n^{\text{ask}}]-c)$. Since $\lambda c$ is
constant in $\pi$, the policy gradient sees only a per-episode reward that \emph{prices} each ask,
\begin{equation}
  R_i = \text{win}_i - \lambda\, n^{\text{ask}}_i,
\end{equation}
which we optimize with GRPO~\citep{grpo2024}. The price $\lambda$ is not chosen: it is the
\emph{shadow price} of an ask (in units of win-probability), solved for by dual ascent until
the budget is met,
\begin{equation}
  \lambda \leftarrow \max\!\big(0,\; \lambda + \eta\,(\overline{n^{\text{ask}}} - c)\big),
\end{equation}
raising the price when the mean ask-rate $\overline{n^{\text{ask}}}$ exceeds $c$ and lowering it
otherwise, with dual step size $\eta$. Note the budget never enters the reward; it acts only
through $\lambda$. A KL anchor to the Stage-2 policy $\pi_{\text{DPO}}$ keeps the update from
eroding learned competence. We choose $c$ with a short sweep over integer budgets that bracket
the ask-rates observed at evaluation (Appendix~\ref{app:frontier}).

\section{Experimental Setup}
\label{sec:setup}

\subsection{Datasets and harnesses}
\label{sec:datasets}

We evaluate on three benchmarks (Table~\ref{tab:datasets}, Appendix~\ref{app:data}), each chosen to stress-test the
advice-seeking policy under a different agentic capability: long-horizon planning, code
generation, and deep web search and retrieval. Preference pairs and reinforcement-learning
rollouts are drawn only from the training splits, and the construction of each benchmark, its
splits, and our training hyperparameters are described in Appendix~\ref{app:data}.

\paragraph{Models.}
To motivate our problem statement, we benchmark two open-weight small models
(Qwen3-8B~\citep{qwen3}, Llama-3.1-8B) and three closed-source \textit{frontier small/lite}-scale
models (Gemini 3.5 Flash-Lite, GPT-5.4 nano, Claude Haiku 4.5), whose sizes are undisclosed but
which let us test whether the advisor--executor pattern works out of the box for any small or
lite model family. We then adopt the better-performing Qwen3-8B as our policy and use Gemini 3.7
Flash as the advisor throughout, evaluating robustness to other advisors in
Section~\ref{sec:results}.

\paragraph{Metrics.}
\textbf{SR} (success rate), our primary measure of quality, is the fraction of test episodes
the policy solves, judged by the environment's own signal. Two quantities describe \emph{how much}
the policy asks: \textbf{asks/ep}, the mean number of advisor calls per episode, and \textbf{\%
eps ask}, the fraction of episodes with at least one call, which together separate a policy that
asks a little on many episodes from one that asks a lot on a few. \textbf{cost/ep} is the mean
advisor spend per episode, which tracks asks/ep but also reflects that advice length varies
across domains; our GRPO stage (Section~\ref{sec:grpo}) budgets asks/ep rather than token cost,
since the former is a definite count tied to the steps of an episode.

\subsection{Baselines}
\label{sec:baselines}

\paragraph{Prompting baselines.} We first evaluate the policy paired with the advisor under two
prompting conditions. \textbf{Neutral} defines \texttt{ask\_advisor} only through its tool
description and argument usage, as any ordinary tool would be, while \textbf{Nudged}
additionally describes in the system prompt, at a high level, the situations in which the policy
should consider asking, giving an upper bound on prompt-level gains (example prompts for
TextWorld are in Appendix~\ref{app:prompts}).

\paragraph{Post-training baselines.} We first compare against an SFT (supervised fine-tuning)
baseline, \textbf{RFT} (rejection-sampling fine-tuning)~\citep{rft2023}, where we fine-tune on
rollouts generated by the base policy paired with an advisor and rejection-sample for
successful episodes using $K=8$ generations per training task. These rollouts therefore contain
organic instances of both advisor use and solo, advisor-free episodes. We further apply
post-hoc perturbations (dropped observations, faulty tool outputs, altered inputs) to the
sampled rollouts, as suggested in R2V-Agent~\citep{r2v2026}, for more robust training. The
checkpoint obtained from this baseline serves as the warm-start SFT stage of our pipeline,
$\pi_{\text{SFT}}$.

While RFT provides sparse feedback over complete traces, \textbf{Step-DPO}~\citep{stepdpo2024}
gives fine-grained feedback at individual steps through DPO. We train it on the same three sets of
preference pairs constructed in Stage~2 (Section~\ref{sec:mining}), using one-step updates for all
of them, so it differs from our Stage~2 only on Follow pairs, where we backpropagate over the whole
continuation rather than the single next step.

We also evaluate against an \textbf{RL GRPO baseline} from the AgentRL~\citep{agentrl2025}
framework, warm-started from the same SFT checkpoint and optimizing the reward
$R=\mathbb{1}[\text{win}]-\lambda\,\#\text{asks}$, where the fixed ask price $\lambda$ is chosen
by a three-point sweep whose bounds we estimate from the distribution of
$P_{\text{ask}}-P_{\text{noask}}$, the \textit{marginal value of asking}, in our counterfactual
replay sweep.

\paragraph{Upper bounds.} \textbf{Oracle} is a hindsight diagnostic: at each candidate step it
takes the ask-or-act decision that our counterfactual replay (Section~\ref{sec:mining}) shows
to be best, so it upper-bounds any timing policy given foresight of episode success obtained
through rollout verification. \textbf{Advisor-only} runs the advisor as a solo policy on each
task and bounds the quality attainable if every decision were delegated to the stronger model.

\section{Results}
\label{sec:results}

Table~\ref{tab:main} reports our main results across the three benchmarks.

\begin{table}[t]
  \centering
  \caption{Main results on the held-out test split of each benchmark
  (Table~\ref{tab:datasets}, Appendix~\ref{app:data}). Policy Qwen3-8B, advisor Gemini~3.7~Flash, thinking on. Per
  benchmark we report success rate (SR), advisor calls per episode (a/ep), and the fraction of
  episodes that ask (\%ask). The top block is the untrained policy under the Neutral and Nudged
  prompts (Appendix~\ref{app:prompts}); the middle block compares the post-training baselines
  and \textbf{ADAPT}; the bottom block gives the non-deployable Oracle and advisor-only upper
  bounds (Section~\ref{sec:baselines}).}
  \label{tab:main}
  \footnotesize
  \setlength{\tabcolsep}{3pt}
  \resizebox{\textwidth}{!}{%
  \begin{tabular}{@{}l ccc ccc ccc@{}}
    \toprule
    & \multicolumn{3}{c}{\textbf{TextWorld}} & \multicolumn{3}{c}{\textbf{LiveCodeBench-stdin}}
    & \multicolumn{3}{c}{\textbf{GAIA-Text}} \\
    \cmidrule(lr){2-4}\cmidrule(lr){5-7}\cmidrule(lr){8-10}
    Method & SR & Ask/ep & \%Ask & SR & Ask/ep & \%Ask & SR & Ask/ep & \%Ask \\
    \midrule
    \multicolumn{10}{@{}l}{\emph{Out-of-the-box (Qwen-3 8B+Advisor)}}\\
    Neutral & 24.5 & 0.10 & 11 & 21.8 & 0.06 & 9 & 17.5 & 0.08 & 9.6 \\
    Nudged  & 28.2 & 0.12 & 13.5 & 35.1 & 0.88 & 64.1 & 19.4 & 0.112 & 12.5 \\
    \midrule
    RFT~\citep{rft2023} (SFT)          & 58.7 & 0.18 & 19.62 & 48.25 & 1.01 & 60 & 36.8 & 0.15 & 13.33 \\
    Step-DPO~\citep{stepdpo2024} (DPO)         & 66.1 & 3.82 & 61.5 & 49.33 & 2.25 & 77.1 & 39.8 & 3.12 & 75.5 \\
    AgentRL~\citep{agentrl2025} (SFT+RL)        & 41.45 & 1.5 & 91 & \textbf{58.3} & 1.15 & 88 & 42.85 & 1.8 & 92 \\
    \textbf{ADAPT (ours)} & \textbf{77.32} & 1.09 & 59 & 56.01 & 1.02 & 66.33 & \textbf{49.25} & 2.12 & 68 \\
    \midrule
    \multicolumn{10}{@{}l}{\emph{Upper bounds}}\\
    Oracle(Qwen-3 8B(base) + Advisor)                     & 86 & 2.8 & 74 & 52 & 1.88 & 81 & 62 & 3.68 & 89 \\
    Advisor only (Gemini 3.7 Flash) & 90.8 & -- & -- & 79.5 & -- & -- & 77.8 & -- & -- \\
    \bottomrule
  \end{tabular}}
\end{table}

\paragraph{ADAPT approaches the oracle ceiling at a calibrated ask-rate.}
ADAPT attains the best success rate on TextWorld ($77.32$) and GAIA-Text ($49.25$) and comes
within two points of the strongest baseline on LiveCodeBench ($56.01$ against $58.3$), while
asking sparingly. Relative to the out-of-the-box Nudged policy, which shares ADAPT's harness and
advisor, the recipe closes $85\%$ of the gap to the Oracle timing ceiling on TextWorld
($28.2\!\to\!77.32$ against an oracle of $86$) and $70\%$ on GAIA ($19.4\!\to\!49.25$ against
$62$). On TextWorld ADAPT reaches its
$77.32$ at only $1.09$ asks per episode, below the oracle's own $2.8$, making it the most
efficient trained policy per advisor call on every benchmark.

\paragraph{No single ingredient suffices; the stages are complementary.}
The three post-training baselines each isolate one stage of our recipe, and each fails in a
characteristic way. Rejection-sampling fine-tuning (RFT) builds solo competence but inherits the
base policy's under-asking: on TextWorld it lifts success to $58.7$ yet asks on only $19.6\%$ of
episodes, leaving the oracle headroom untouched. Step-DPO adds the asking behavior but has no
notion of what an ask costs, and over-asks everywhere. The RL-only baseline (AgentRL) drives ask
coverage up but, absent a warm-started competence to reinforce, collapses on the sparse-reward
long-horizon domains, falling to $41.45$ on TextWorld, below even RFT. ADAPT supplies these competences in sequence, and the combination is best or near-best on all
three benchmarks at an ask-rate none of the isolated baselines match.

\paragraph{Over-asking takes two distinct forms: depth and breadth.}
Separating asks per episode from the fraction of episodes that ask exposes two different
miscalibrations. Step-DPO carries the highest asks per episode among trained methods on every
benchmark ($3.82$, $2.25$, $3.12$) while asking on a moderate fraction of episodes, so when it
decides to ask it asks repeatedly, whereas the RL baseline asks on almost every episode
($88$--$92\%$) but only around once, since the ask penalty attenuates the count while leaving
coverage broad. ADAPT instead modulates its ask-rate by domain, asking about once per episode on
TextWorld and LiveCodeBench but $2.12$ times on the longer, advice-hungry GAIA episodes, tracking
where advice is actually worth its price.

\paragraph{LiveCodeBench inverts the trend, and reveals when advice is not the limiting factor.}
Its episodes are short, roughly four turns on average, and it is the only environment with a
granular verifier, whose unit tests give step-level feedback that rewards a model's innate
self-correction. The RL baseline is the strongest method here ($58.3$) and, together with ADAPT
($56.01$), it exceeds the Oracle ($52$), which runs the base policy with perfectly timed advice;
here the dense test signal lets reinforcement learning improve self-correction more than
well-timed advice can. ADAPT stays competitive by asking sparingly, but this marks the boundary
of the advisor--executor advantage: where the executor can cheaply verify and repair itself,
asking matters less.

\paragraph{Prompting is not enough, and its effect is domain-dependent.}
The Neutral and Nudged conditions both fall far short of training. Nudging moves TextWorld and
GAIA success by only a few points and helps materially only on short-horizon LiveCodeBench
($21.8\!\to\!35.1$), where a single early ask is easy to place, confirming
Table~\ref{tab:motivation} that the miscalibration of the out-of-the-box pattern is not
something prompting repairs.

\paragraph{Robustness to a changing advisor.} Swapping ADAPT's Gemini~3.7~Flash advisor for a
stronger (Claude Opus~4.8) or weaker (DeepSeek V4 Flash) one, both unseen in training, keeps
success well above the untrained policy everywhere: on TextWorld $77.3\%$ rises to $79\%$ with
Opus and dips only to $68\%$ with DeepSeek (Appendix~\ref{app:advisor}), so the learned
help-seeking transfers across the advisor shift rather than fitting one teacher.

\section{Conclusion and Future Work}
\label{sec:conclusion}

Off-the-shelf small models, open and frontier alike, do not know when to ask a stronger model for
help, and we showed that ADAPT can teach them from the policy's own environment-verified
rollouts, improving success across all three domains while keeping advisor calls low and closing
much of the gap to an oracle timing ceiling on the long-horizon tasks.

Two directions stand out. First, our labels come from environment verification of rollouts, which
fits many agentic workloads but not tasks with a more mixed or subjective notion of success; these
will need LLM-as-judge or PRM scoring where a rollout cannot be checked directly. Second, under
limited compute we trained only LoRA adapters on modest data, so scaling to more policy models,
larger datasets, and full fine-tuning remains to be tested.

\bibliography{references}

\appendix

\section{Preference-pair summary}
\label{app:pairs}

Table~\ref{tab:pairs} summarizes the three preference pairs mined in Stage~2
(Section~\ref{sec:mining}), together with the base-rollout outcome each is mined from and the
thresholds that select it.

\begin{table}[h]
  \centering
  \caption{The three preference pairs ADAPT mines from counterfactual replay, with
  the base-rollout outcome each is mined from and the thresholds on $\hat p_{\text{ask}}$,
  $\hat p_{\text{noask}}$, and the ask-advantage $A$ that select it.}
  \label{tab:pairs}
  \small
  \setlength{\tabcolsep}{4pt}
  \begin{tabular}{@{}lllll@{}}
    \toprule
    Pair & Source & Selection criterion & chosen / rejected & Teaches \\
    \midrule
    \textbf{Ask}     & unsolved & $\hat p_{\text{ask}}>0.6,\ A>0.3$   & ask / act & when to ask \\
    \textbf{Abstain} & solved   & $\hat p_{\text{noask}}>0.6,\ A\le 0$ & act / ask & when not to ask \\
    \textbf{Follow}  & both     & $\hat p_{\text{ask}}\in(0,1)$        & advice followed / drifted & how to use advice \\
    \bottomrule
  \end{tabular}
\end{table}

\section{The advisor budget as a cost dial}
\label{app:frontier}

\begin{figure}[t]
  \centering
  \includegraphics[width=0.68\linewidth]{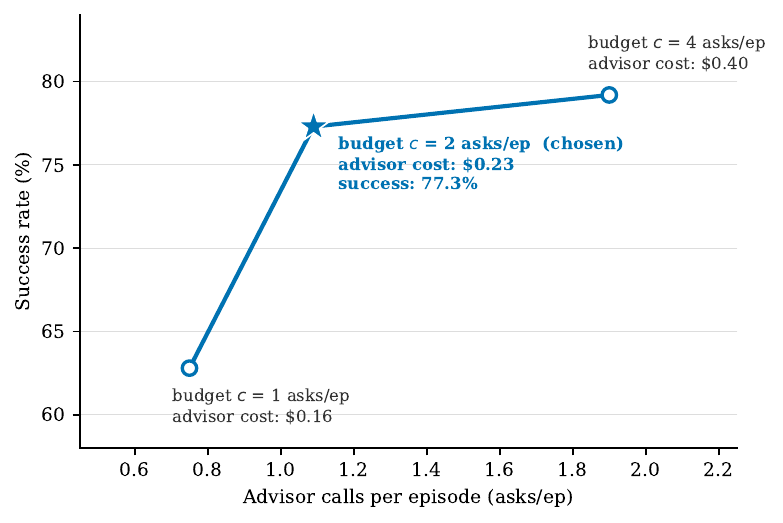}
  \caption{\textbf{The advisor budget as a cost dial (TextWorld).} Success rate against realized
  advisor calls per episode as the Stage-3 budget $c$ (a ceiling on asks/ep) is varied. Advisor
  cost is the spend over the full 109-game evaluation at the measured rate of \$0.00193 per call.
  The chosen operating point ($c=2$) reaches $77.3\%$ at $1.09$ asks/ep for \$0.23; a tighter
  budget ($c=1$) gives up roughly $15$ points of success for lower cost, while a larger budget
  ($c=4$) adds under two points, because extra asks stop converting. Only the $c=2$ point is a measured
  ADAPT result; the
  $c=1$ and $c=4$ points illustrate a budget sweep we leave to future work.}
  \label{fig:frontier}
\end{figure}

The Stage-3 budget $c$ (Section~\ref{sec:grpo}) is not tuned to an optimum; it is an operating
point the practitioner chooses by cost tolerance. Sweeping $c$ traces a success-versus-cost
trade-off (Figure~\ref{fig:frontier} for TextWorld): raising the budget lets the policy ask more,
with diminishing returns, and the deployment choice is simply where on this curve to sit.
On TextWorld we use $c=2$, under which the trained policy settles at an observed $1.09$ asks/ep,
well under the ceiling, and we set $c$ the same way on the other benchmarks; full values are in
the code.

\section{Benchmark and training details}
\label{app:data}

\begin{table}[h]
  \centering
  \caption{The three benchmarks and their harnesses. Tools are harness-specific;
  \texttt{ask\_advisor} (Section~\ref{sec:interface}) is available in every harness.
  \emph{Hor.}\ is the environment-step budget; for GAIA-Text it is the main-agent loop, whose web and file sub-agents run $20$ steps each.}
  \label{tab:datasets}
  \footnotesize
  \setlength{\tabcolsep}{2.5pt}
  \resizebox{\textwidth}{!}{%
  \begin{tabular}{@{}l l >{\raggedright\arraybackslash}p{2.6cm} >{\raggedright\arraybackslash}p{3.3cm} l c c@{}}
    \toprule
    Benchmark & Domain & Harness & Tools & Success & Hor. & Train / Test \\
    \midrule
    TextWorld           & embodied control & AgentBench-FC~\citep{agentbench2024} & \texttt{take\_action}, \texttt{ask\_advisor} & win flag & 20 & 600 / 109 \\
    LiveCodeBench-stdin & code generation  & ours, cf.\ R2V~\citep{r2v2026}       & \texttt{write\_code}, \texttt{run\_tests}, \texttt{submit}, \texttt{ask\_advisor} & tests pass & 10 & 323 / 134 \\
    GAIA-Text           & deep web search  & CK-Pro~\citep{ckpro2025}             & \texttt{search}, \texttt{browse}, \texttt{ask\_advisor} & Exact match & 15 & 680 / 103 \\
    \bottomrule
  \end{tabular}}
\end{table}

\textbf{TextWorld} is built on ALFWorld~\citep{alfworld2021}, a
household-control benchmark where the agent issues textual commands such as
\texttt{go to}, \texttt{take}, \texttt{put}, \texttt{clean}, and \texttt{heat} across six
embodied task families. We use the AgentBench~\citep{agentbench2024} framework, which hosts
several environments including ALFWorld, and build our training split by randomly sampling
$600$ episodes (seed $42$) from the $3{,}163$ tasks in its train split, evaluating on the
standard test split of $109$ tasks.

\textbf{LiveCodeBench-stdin} draws on the standard-I/O subset of the broader LiveCodeBench
benchmark~\citep{livecodebench2024}, which we model as a multi-step coding environment whose
actions are \texttt{write\_code}, \texttt{run\_tests}, and \texttt{submit}, following the R2V
setup~\citep{r2v2026}. From a pool of $498$ problems we use $457$ in total: the test set
consists of $134$ medium and hard problems, and to guard against contamination we retain only
problems released after the Qwen3-8B training cutoff
(\texttt{contest\_date}\,$\ge$\,2024-08-01).

\textbf{GAIA-Text}~\citep{gaia2023} is the text-input-only $103$-question benchmark introduced
by WebThinker~\citep{webthinker2025}, for which we use the Cognitive
Kernel-Pro~\citep{ckpro2025} framework as our deep-research harness. Since GAIA provides no
training split, we follow CK-Pro and train on the open-source WebWalkerQA subset of $680$
queries.\footnote{\url{https://huggingface.co/datasets/CognitiveKernel/CognitiveKernel-Pro-Query}}

Across all three benchmarks, preference pairs and reinforcement-learning rollouts are drawn
\emph{only} from the training split, and we hold out an $85/15$ validation slice of each
training split for hyperparameter tuning. Our test splits are standard wherever one is
defined, and we provide our full training and test splits alongside our code.

\paragraph{Training setup.}
All three stages adapt the policy with LoRA~\citep{lora2022} rather than full fine-tuning: rank
$32$, $\alpha=64$, applied to all linear projections, with a maximum sequence length of
$16{,}384$ tokens to fit the long reasoning traces. The supervised warm-start (Stage~1) is
trained at learning rate $3\times10^{-4}$; Stage~2 runs DPO for $2$ epochs at learning rate
$1\times10^{-5}$ with $\beta=0.1$, taking the frozen Stage-1 checkpoint as the reference;
Stage~3 runs cost-constrained GRPO at learning rate $1\times10^{-5}$ with a PPO-style ratio
clip of $0.2$ and a KL anchor to $\pi_{\mathrm{DPO}}$ to preserve the competence already
learned. We release our code and training splits for reproducibility.

\section{Robustness to a changing advisor}
\label{app:advisor}

The advisor available at deployment is rarely the one used in training. We evaluate the trained
ADAPT policy (Qwen3-8B, trained with Gemini~3.7~Flash as advisor) with two advisors it never saw
during training: a weaker DeepSeek Flash and a stronger Claude Opus~4.8. Table~\ref{tab:advisor}
reports success rate and advisor calls per episode on all three benchmarks, together with each
advisor's solo (advisor-only) performance for reference. The trained policy retains most of its
gains under both swaps, indicating that the learned help-seeking is a property of the policy
rather than of the specific advisor it was trained with.

\begin{table}[h]
  \centering
  \caption{\textbf{Advisor swap.} The ADAPT policy (Qwen3-8B, trained with Gemini~3.7~Flash)
  evaluated with a weaker and a stronger advisor unseen in training. We report success rate (SR)
  and advisor calls per episode (a/ep) per benchmark; the lower block gives each advisor's
  solo performance for reference. Gemini~3.7~Flash rows are the deployed configuration from
  Table~\ref{tab:main}. The DeepSeek V4 Flash and Claude Opus~4.8 rows are single-seed
  evaluations, whereas the main results average three seeds.}
  \label{tab:advisor}
  \footnotesize
  \setlength{\tabcolsep}{5pt}
  \begin{tabular}{@{}l cc cc cc@{}}
    \toprule
    & \multicolumn{2}{c}{\textbf{TextWorld}} & \multicolumn{2}{c}{\textbf{LiveCodeBench-stdin}}
    & \multicolumn{2}{c}{\textbf{GAIA-Text}} \\
    \cmidrule(lr){2-3}\cmidrule(lr){4-5}\cmidrule(lr){6-7}
    Advisor & SR & a/ep & SR & a/ep & SR & a/ep \\
    \midrule
    \multicolumn{7}{@{}l}{\emph{ADAPT policy (Qwen3-8B) $+$ advisor}}\\
    Gemini 3.7 Flash (train)      & 77.3 & 1.09 & 56.0 & 1.02 & 49.3 & 2.12 \\
    DeepSeek V4 Flash             & 68   & 1.45 & 53   & 1.15 & 42   & 2.4  \\
    Claude Opus 4.8               & 79   & 1.1  & 62   & 1.1  & 57   & 2.3  \\
    \midrule
    \multicolumn{7}{@{}l}{\emph{Advisor only (solo)}}\\
    Gemini 3.7 Flash              & 90.8 & --   & 79.5 & --   & 77.8 & --   \\
    DeepSeek V4 Flash             & 78   & --   & 84   & --   & 76   & --   \\
    Claude Opus 4.8               & 98   & --   & 94   & --   & 89   & --   \\
    \bottomrule
  \end{tabular}
\end{table}

\section{Prompt conditions}
\label{app:prompts}

The two prompting baselines (Section~\ref{sec:baselines}) rewrite only the tool clause of the
stock system prompt; the environment never sees the change, and the \texttt{ask\_advisor} tool
description is identical in both. We show the TextWorld prompts below.

\paragraph{\texttt{ask\_advisor} tool description (both conditions).} ``Consult an expert
advisor about your current task. The advisor can see your task and your full progress so far,
and replies with strategic guidance on how to approach and solve the task from where you are
now---an overall plan and what to prioritize, not a specific next command. Available at any
point in the episode.'' Its single argument, \texttt{request}, is described as ``Briefly, what
you want help with or are trying to figure out.''

\paragraph{Neutral.} ``A \texttt{take\_action} tool will be provided for you to submit the
action you want to take in the environment. You also have an \texttt{ask\_advisor} tool: it
consults a stronger expert advisor that can see your task and your full progress and returns
strategic guidance on how to proceed.''

\paragraph{Nudged.} The Neutral text above, followed by: ``Consider using it when you are
unsure how to approach the task, when your actions keep failing or you seem to be making no
progress, or before you conclude the task---then continue acting yourself, guided by its
advice.''

\end{document}